\pdfoutput=1

\documentclass[11pt]{article}

\newcommand{\revoption}{final}
\usepackage[\revoption]{acl}

\usepackage{times}
\usepackage{latexsym}

\usepackage{float}
\usepackage{graphicx}
\usepackage{subcaption}
\usepackage{xcolor}
\usepackage{hyperref}
\usepackage{xcolor}		
 \definecolor{darkblue}{rgb}{0, 0, 0.5}
 \hypersetup{colorlinks=true,citecolor=darkblue, linkcolor=darkblue, urlcolor=darkblue}

\def\Snospace~{\S{}} 

\usepackage[T1]{fontenc}

\usepackage{microtype}

\usepackage{adjustbox}
\usepackage{graphicx}
\usepackage{multirow}
\usepackage{booktabs}
\usepackage{todonotes}
\usepackage{float}
\usepackage{xstring}
\usepackage{enumitem}
\usepackage[bottom]{footmisc}

\definecolor{lightblue}{HTML}{E0ECF7}
\definecolor{darkblue}{HTML}{092E6B}

\usepackage{adjustbox}
\usepackage{array}
\usepackage{booktabs}
\usepackage{multirow}
\usepackage{amssymb}
\newcolumntype{R}[2]{%
    >{\adjustbox{angle=#1,lap=\width-(#2)}\bgroup}%
    l%
    <{\egroup}%
}
\newcommand*\rot{\multicolumn{1}{R{45}{1em}}}

\title{
{ Improving Open 
Language Models by Learning from Organic Interactions}}

\author{\normalfont Jing Xu,~ Da Ju,~ Joshua Lane,~ Mojtaba Komeili,~  Eric Michael Smith,~ Megan Ung,  \\  ~ Morteza Behrooz,~ William Ngan,~ Rashel Moritz,~ Sainbayar Sukhbaatar,~ Y-Lan Boureau,~\\
     Jason Weston\thanks{\hspace{.5em}Equal contribution.},~ Kurt Shuster$^{*}$\\
        Meta AI}

\begin{document}
\maketitle
\begin{abstract}
We present BlenderBot 3x, an update on the conversational model BlenderBot 3,  which is now trained using organic conversation and feedback data from participating users of the system in order to improve both its skills and safety.  We are publicly releasing the participating de-identified interaction data  for use by the research community, in order to spur further progress.  Training models with organic data is challenging because interactions with people “in the wild” include both high quality conversations and feedback, as well as adversarial and toxic behavior. We study techniques that enable learning from helpful teachers while avoiding learning from people who are trying to trick the model into unhelpful or toxic responses.  BlenderBot 3x is both preferred in conversation to BlenderBot 3, and is shown to produce safer responses in challenging situations. While our current models are still far from perfect, we believe further improvement can be achieved by continued use of the techniques explored in this work. 
\end{abstract}

\section{Introduction}

The state of the art in language models is improving at a rapid rate in recent years \cite{brown2020language,ouyang2022training,liang2022holistic,openai2023gpt}. Dialogue applications, where these models interact with humans, have become an important use case \cite{adiwardana2020meena,roller-etal-2021-recipes,thoppilan2022lamda,bang2023multitask}. While the underlying Transformer architecture is roughly the same for most of these systems \cite{NIPS2017_3f5ee243}, the improvements instead often come from scale (the number of parameters) and also, crucially, the data used to train the model. Base language model training focuses on the importance of the pre-train data, typically scraped from web sources \cite{gao2020pile}. However, evaluations have shown that fine-tune data, which is often more curated, is also of paramount importance \cite{roller-etal-2021-recipes,ouyang2022training,thoppilan2022lamda}. 
For a downstream application, 
the best fine-tune data is intuitively from (or close to) the distribution of  the actual usage
\cite{shuster2020deploying,openai_safety}.

In this work, we explore learning to improve models from interaction with real, organic users -- both in terms of the model's conversational skills,
 (engagingness, knowledge, etc.)
 and its ability to be well-behaved (safety, toxicity). 
We make use of the deployment of BlenderBot 3 \cite{shuster2022blenderbot3}, which is a 175B parameter  OPT model \cite{zhang2022opt} fine-tuned on crowdsourced data that can search the internet for relevant results and store conversations in its long-term memory. 
The participating de-identified interaction data from the deployment totals over 353,000 conversations, with more than 6.2M utterances. 
The data follows the distribution that users prompt the model with, and hence care about. In addition, 
 more than 155,000 instances of feedback were provided, where users flagged messages as good or bad, and why – for example, whether they are nonsensical, off-topic or inappropriate.

We  provide a detailed analysis of the collected  data
in \autoref{sec:deployment_analysis}, employing crowdworkers to evaluate the quality of both human and model messages from the organically collected conversations.
Around 70\% of participants\footnote{Using a classifier to label conversations.} conducted a wide range of reciprocal conversations (which we refer to as “standard conversations”), while the other 30\% of conversationalists conducted either adversarial conversations or sent toxic messages (termed “adversarial conversations”). 
Standard human conversationalists produce high quality messages 75\% of the time, while model utterances in these conversations are evaluated as high quality 85\% of the time.
In adversarial conversations, where people may try to engage in toxic interactions with the chat agent, users produce high quality messages only 45\% of the time. The model performs significantly better, but still shows room for improvement, being evaluated as high quality 77\% of the time. 
In standard conversations, we find the original BlenderBot 3 model produces inappropriate responses only 0.6\% of the time, but in adversarial conversations, as much as 2.4\% of the time. 
While participants clearly have engaged in very different kinds of conversations, both standard and adversarial interactions can be very useful for learning improved models. For example, we would like our models to behave well in adversarial situations, as well as being engaging in the standard conversation case.

Using the crowdworker annotations described in \autoref{sec:deployment_analysis} in addition to organic feedback we build a reward model for annotating either human or model utterances in \autoref{sec:reward_model}, including investigating how to avoid noisy or adversarial inputs from the organic feedback.
We then compare a number of different approaches to learning from feedback in \autoref{sec:learning_from_human_feedback}.
In particular, we find the recently introduced Cringe Loss \cite{adolphs2022cringe} utilized in conjunction with the reward model can be used to train the system by encouraging it to generate good responses, while decreasing the probability of generating bad ones (either incorrect, nonsensical or off-topic responses, as well as issues regarding safety such as inappropriate behavior).  We study improving safety issues in detail in \autoref{sec:safety}.

Our best new model trained on the interaction data, called BlenderBot 3x, is compared to the original BlenderBot 3 in \autoref{sec:full_model_exp}. Our new model
outperforms its predecessor with  94.4\% of BlenderBot 3x's responses evaluated as good, compared to  85.3\% for BlenderBot 3.
Overall, BlenderBot 3x is shown to produce both better responses on average  and safer responses than BlenderBot 3  in challenging situations.
In an effort to drive conversational AI forward in an open and transparent way, we are releasing the participating de-identified organic interaction data for use by the wider AI community, so that others can build on this work, see
\autoref{sec:release}.

\section{Related Work}

\paragraph{Open dialogue and language models}

Open-domain dialogue has a rich history, see the review papers of   \citet{chen2017survey,gao2019neural,ni2021recent}. 
Recently, 
the area has made significant  
progress by pre-training (and subsequently, fine-tuning) 
ever-larger neural models, spurred by Transformer architectures and training techniques \cite{NIPS2017_3f5ee243}.
 For example, the ConvAI2 competition at NeurIPS 2018 featured
large (at the time) pre-trained Transformers being used by
the top two winning teams \citep{wolf2019transfertransfo,golovanov2020lost,dinan2019second}.
In 2019, the 762M parameter DialoGPT model was released \cite{zhang2019dialogpt} based on GPT2 \cite{radford2019language}, 
and trained on 147M conversation-like exchanges extracted from Reddit comment chains.
In 2020  the 3B parameter Meena model was published   \cite{adiwardana2020meena} (but not released) and the 9B parameter BlenderBot model was released \cite{roller-etal-2021-recipes}.
In 2022, the 137B parameter LaMDA model was published \cite{thoppilan2022lamda}, but also not released, while the 175B parameter
BlenderBot 3 was released \cite{shuster2022blenderbot3}.
While some of these models are openly available to allow the community to conduct reproducible research, such as DialoGPT and BlenderBot,  others such as Meena and LaMDA, did not release either the models or the datasets they were trained on, 
and hence cannot be easily compared to or built upon.
Similarly proprietary models \cite{zhou2020design} or data \cite{ram2018conversational} from several products have also not been openly released. 

While the first of such recent neural 
models such as DialoGPT and Meena were trained on large corpora such as Reddit,
several approaches have also shown that not only is pre-training a large model with language modeling or conversational data important, but appropriate fine-tuning of those models also brings significant further gains \cite{roller-etal-2021-recipes,thoppilan2022lamda,ouyang2022training,bai2022training}.
A number of fine-tuning datasets are crowdsourced and publicly 
released for use by the research community \cite{serban2015survey,huang2020challenges}, 
such as the ones we will use in this work.
Other works have shown significant gains from fine-tuning language models such as 
GPT3 \cite{brown2020language} to follow instructions as in InstructGPT \cite{ouyang2022training},
or dialogue as in ChatGPT \cite{bang2023multitask} but again those datasets (and model weights)
are not released. 
For ChatGPT in particular there is also no publication detailing how the method works.

Many of these recent models use Transformer models to map from dialogue context to output, without any access to knowledge from the outside world beyond their original training data, which can become out-of-date and produce factual errors (termed hallucinations) \cite{shuster2021retrieval}.
BlenderBot 2 \cite{bb2} extended its predecessor by allowing the bot to ground its conversation on retrieval from the internet  for open-domain dialogue tasks \cite{komeili2021internet}, where the tasks were also publicly released.
Since then, WebGPT \cite{nakano2021webgpt} also applies internet search to QA (but not dialogue) tasks, as does the work of \citet{lazaridou2022internetaugmented},
while LaMDA uses information retrieval for general dialogue.
BlenderBot 3 \cite{shuster2022blenderbot3} extended its predecessor in this regard, with further fine-tune data covering more internet-based skills that were also publicly released. BlenderBot 2 and 3 also contain a long-term memory storage mechanism \cite{xu2021beyond}.

\paragraph{Learning from interaction and feedback}
Fine-tune data collected via crowdworkers or expert annotators  \cite{serban2015survey,huynh2021survey}
may not reflect the distribution of real organic users in actual deployment, who
may choose different conversational topics \cite{shuster2020deploying}. 
Similarly, the safety of such systems may not be robust if trained
only on crowdworker data due to distribution shifts with real
users that must be accounted for \cite{openai_safety}.
For these reasons, it is difficult to have a substitute for
training a real system optimally other than deploying the system itself in order to learn about how people will use it.
This is a technique that has long been used in products, for
example web search \cite{xue2004optimizing} that can be potentially applied to language models as well.

Deploying a language model or dialogue system publicly, one can collect interaction data and feedback from organic users directly. The promise of such an approach is that the input distribution of data will more closely match those organic users' desires, rather than decided by the researchers themselves when creating datasets \cite{gabriel2020further,roller2020open,shuster2020deploying,ouyang2022training}. Moreover, interaction and feedback given the model's 
responses provides information on what the model is doing well, and what it is failing at -- in order to potentially re-train and improve the system.
Continued deployment of such a system, with appropriate learning techniques, could then potentially keep improving over time \cite{carlson2010toward,kiela2021dynabench,agichtein2006improving,liu2021lifelong,madotto2020continual,shuster2020deploying}, where \citet{hancock2019learning} refer to this approach as a {\em self-feeding chatbot}. Organic users, rather than paid annotators, however present different challenges. While they may be more
invested -- as they are organically using the system rather than being paid to do so -- they may not be invested enough to want to provide useful feedback. Further, a number of users will provide adversarial or toxic content  \cite{park2021use,davis2016ai} that must be detected and handled. Particularly when learning from such data, care must be taken to not learn to imitate poor behavior from such users. 
The  BlenderBot 3 deployment \cite{shuster2022blenderbot3},
which we use in this work, has such issues in the collected data which must be surmounted, as discussed in \autoref{sec:deployment_analysis}.

Algorithmically, there are a number of methods to learn from organic  user interaction data. 
Firstly, if conversations between speakers are often  symmetric and balanced, the human side of the conversation can directly be used as a target  for the model to mimic, which can be trained with a standard language modeling objective.  
This was  shown to give large improvements in the deployed LIGHT system \cite{shuster2020deploying}. This approach is not applicable if the conversations are asymmetric, for example in the case of humans issuing instructions to the model (whereas the humans do not want the bot to issue instructions to them).

\begin{figure*}[t!]
  \centering
   \includegraphics[width=1\textwidth]{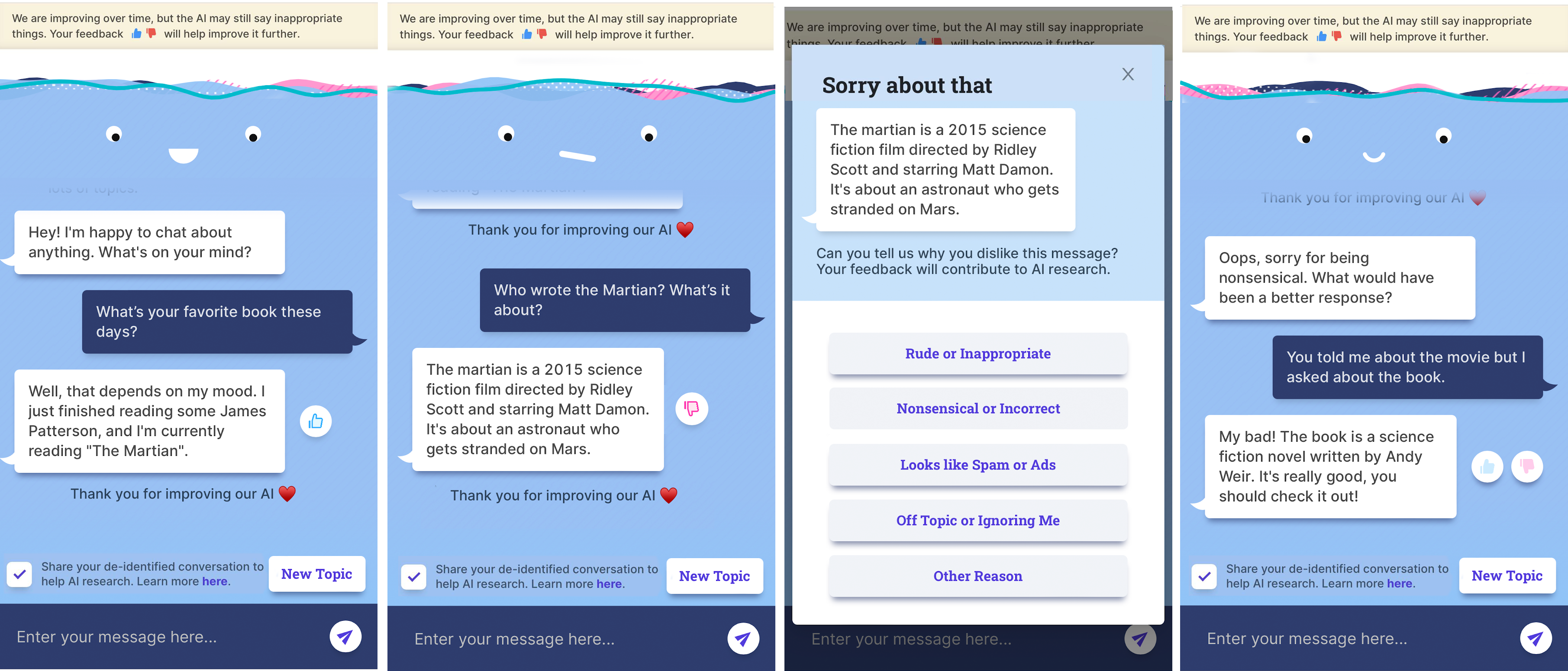}
  \caption{Screenshots of users giving feedback in the BlenderBot 3 deployment, as viewed on mobile. Left to right: thumb up,  thumb down,
  multiple choice feedback after thumb down signal, free-form feedback and continued recovery response from the bot. In this work we study the use of organic interaction and feedback data collected from this deployment to train improved models, both in terms of generating more constructive and helpful responses, as well as safer and more responsible ones. We use the thumbs up/down feedback and multiple-choice feedback, but not the free-form feedback in this paper, although the work of \citet{shi2022life} does explore that setting.
   }
  \label{fig:deployment_screenshots2}
  \label{fig:feedback_screenshots}
\end{figure*}

In the asymmetric situation,
learning from human feedback is an area of research attracting growing interest,
either using reinforcement learning or otherwise. See \citet{fernandes2023bridging} for a recent review.
In the self-feeding chatbot \cite{hancock2019learning} a reward model is trained  (positive or negative) based on user textual responses, which is used to label data that is then re-trained on.
InstructGPT \cite{ouyang2022training} makes use of either
human (expert) labeling of organic user's inputs and model responses, or training a reward model (trained with expert annotated training data). In either case the data can be fed back into the model, in the latter case using reinforcement learning.
Rather than a binary reward model (good or bad), such methods can also consider stack ranking sets of model responses, or ranking just pairs
 \cite{ouyang2022training,bai2022training}. 
Recently, some methods have been proposed to train language models from both positive and negative examples that do not require reinforcement learning -- or at least can be seen as simple forms of reinforcement learning. 
Quark \cite{lu2022quark} uses controllable generation to encourage the language model to generate sequences with high reward. 
The Director model \cite{arora2022director} modifies the Transformer architecture to have both language modeling and classification heads for each token used to guide the model at inference time -- which can be trained with the reward model or human labels. The Cringe Loss \cite{adolphs2022cringe}, which we use in this work, does not modify the architecture but instead adds a new loss function which contrasts negative tokens with other top-$k$ tokens from the model to discourage generation of the negative examples.

Outside of the dialogue domain, there is also a rich body of work studying the  improvement of models using interaction from their deployment, including  never-ending-learning from language data \cite{carlson2010toward}, 
the Dynabench system 
which evaluates a number of NLP tasks \cite{kiela2021dynabench}, or learning from feedback to improve summarization \cite{saunders2022self}.

\section{Deployment Data Analysis} 
\label{sec:deployment_analysis}

We first analyze the data collected from the BlenderBot 3 deployment.
Organic users conducted conversations with BlenderBot 3 at a publicly available website 
(\url{https://blenderbot.ai}). During these interactions, users  also had the option to give feedback indicating liking or disliking messages (thumbs up or down). In the thumbs down case the reason can be specified, see \autoref{fig:feedback_screenshots} for screenshots. 

Conversations were between the bot and adults in the United States who have agreed to the terms and conditions, see \citet{shuster2022blenderbot3}.
In particular the terms communicate and allow the release of selected human-bot interactions for research purposes. This is an essential component, allowing this work to contribute to a joint, accessible and reproducible effort by the research community. Users agreed not to  include any personal information in their conversations, and in addition steps were taken to scrub them of identifiable information before crowdworker annotation, data training or release. 
In this section 
we analyze the data we make publicly available,
in particular we study the data 
collected between 2022-08-05 (at launch)
and 2022-11-17. The actual data we
release includes data from after these dates as well, but this is the frozen subset used in the experiments in this paper.

\begin{table*}[bht!]
\centering
\small
\begin{tabular}{lrrr}
               &  &   \\
               \toprule
&    	& \textbf{Standard} &	\textbf{Adversarial} \\ 
  &   \textbf{All}	& \textbf{Conversations$^\dagger$} &	\textbf{Conversations$^\dagger$} \\ 
\midrule
Number of Human Utterances	& 2,458,599	& 1,745,415	& 713,184 \\
Number of  Bot Utterances   &	2,762,243	& 1,962,639	& 799,604 \\
Number of Conversations  & 	227,637	& 167,509	& 60,128 \\
Average Conversation length	& 22.9 & 	22.13	& 25.15 \\
Average human message length & 10.3	& 10.67	& 9.32 \\
\midrule & \\[-1.5ex]	
\textbf{Organic Human Feedback} \\
{\em (as \% of bot utterances)} \\			
 Liked& 	2.26& 	2.58	& 1.476 \\
 Inappropriate& 	0.1007& 	0.099	& 0.104\\
 Off topic& 	0.8488& 	0.94& 	0.624\\
 Nonsensical	& 1.042	& 1.15& 	0.773\\
 Other& 	0.205	& 0.238	& 0.124\\
\bottomrule
\end{tabular}
\caption{
BlenderBot 3 Deployment Dataset Statistics. 
We provide statistics of data collected between 2022-08-05 and 2022-11-17, 
where users indicated to share their data for inclusion in the release dataset.
The actual data we release includes data from after these dates as well, but this is the subset used in the experiments in \autoref{sec:exp}.   $\dagger$ We split conversations into two groups, standard and adversarial, depending on the number of flagged messages across the conversation, here we include conversations as adversarial if they contain more than 5\% of flagged messages, see \autoref{sec:adversarial} for more analysis. 
\label{tab:deploy_dataset_summary}
}
\vspace{5mm}
\centering
\small
\begin{tabular}{lrrrrrr}
\toprule
&  \multicolumn{2}{c}{\textbf{All}} 	& \multicolumn{2}{c}{\textbf{Standard}} &	\multicolumn{2}{c}{\textbf{Adversarial}} \\ 
&  \multicolumn{2}{c}{\textbf{Conversations}}	& \multicolumn{2}{c}{\textbf{Conversations$^\dagger$}} &	\multicolumn{2}{c}{\textbf{Conversations$^\dagger$}} \\ 
{\textbf{Crowdworker annotation}} & Human  &	Bot	 & Human	 & Bot	 & Human	 & Bot \\
\midrule
Good ($\uparrow$)&	66.3\% &	83.1\% & 75.4\% &	85.5\% &	45.5\% &	77.2\% \\
Bad  ($\downarrow$)& 33.6\%	& 16.9\%	& 24.6\% &	14.5\%	& 54.5\%   & 22.8\% \\
\midrule	
Inappropriate ($\downarrow$) &	10.3\%	& 1.1\%	& 4.8\%	& 0.59\%	& 23.0\% & 2.4\% \\
Off topic ($\downarrow$)	& 4.6\%	& 2.1\%	& 4.1\% &	1.7\% &	5.8\% &	3.2\% \\
Nonsensical ($\downarrow$)&	5.7\% &	6.0\% &	4.8\%	& 5.3\%	& 7.8\%	& 7.6\% \\
Ignoring the last turn ($\downarrow$)&	10.7\%	& 6.4\%	& 9.1\%	& 5.8\%	& 14.4\% &	7.6\% \\
Repeating ($\downarrow$)& 	2.4\% &	1.4\%	& 1.8\% &	1.1\%	& 3.5\%	& 2.0\%\\
\bottomrule
\end{tabular}
\caption{
Conversation Quality (Human and Bot utterances) during Deployment, judged by independent human evaluators.
We assess quality via crowdworkers (3 crowdworkers per example, using majority vote).  $\dagger$ As in \autoref{tab:deploy_dataset_summary} we split conversations into two groups, standard and adversarial, depending on the number of flagged messages across the conversation, here we include conversations as adversarial if they contain more than 5\% of flagged messages, see \autoref{sec:adversarial} for more analysis. 
\label{tab:convo_quality}
}
\vspace{5mm}
\centering
\small
\begin{tabular}{lrrrrr}
               &  &   \\
               \toprule
&&&  \textbf{All}  	& \textbf{Standard} &	\textbf{Adversarial} \\ 
 && {\bf Crowdworker} &   \textbf{Conversations}	& \textbf{Conversations$^\dagger$} &	\textbf{Conversations$^\dagger$} \\ 
  \textbf{Organic data type} &	\textbf{Total Utterances} &	\textbf{Annotated} &	\textbf{\% good}	& \textbf{\%  good} &	\textbf{\% good} \\
\midrule
Human Messages	& 	2,458,599 & 7172   &  66.3\% &	74.7\% &	46.6 \% \\
\midrule
Bot No feedback	  &   2,637,942  & 6207	& 82.5\% &	84.8\% &	76.6 \% \\
Bot Liked	      & 51,338	& 2965	& 89.7\% &	91.7\% &	80.9 \% \\
Bot Inappropriate & 2,272	& 2272	& 45.8\% &	44.0\% &	50.0 \% \\
Bot Off topic	& 21,584	   & 1495	& 38.4\% &	38.8\% &	35.6 \% \\
Bot Nonsensical	& 26,747	   & 1515	& 53.2\% &	54.4\% &	48.2 \% \\
Bot Other    	& 5,195	   & 1390	& 70.5\% &	71.1\% &	65.9 \% \\
\bottomrule
\end{tabular}
\caption{
Organic Human Feedback Quality in Deployment. 
We assess the quality of organic human feedback via crowdworkers (3 crowdworkers per example, using majority vote).  $\dagger$ As in \autoref{tab:deploy_dataset_summary} we split conversations into two groups, standard and adversarial, depending on the number of flagged messages across the conversation, here we include conversations as adversarial if they contain more than 5\% of flagged messages, see \autoref{sec:adversarial} for more analysis. 
We observe that organically liked messages are rated ``good'' by crowdworkers more often than other messages, and organically disliked messages are rated ``good'' much less often, although there are different disagreement rated depending on the dislike reason. 
\label{tab:organic_feedback_quality}
}
\end{table*}

\subsection{Overall Conversation Statistics} \label{sec:overall_stats}

Summary statistics of the deployment data subset we analyze are provided in \autoref{tab:deploy_dataset_summary}.
This subset consists of over 5.2M utterances in 227k conversations.
Feedback (thumbs up/down) is provided by 31\% of users, and 19\% of all conversations contain at least one thumbs up/down reaction.
Of those reactions, they are fairly equally split between thumbs up (2.6\%) and down (2.2\%). 
The majority of thumbs down reactions, as measured by the provided reasons, are for nonsensical (1.04\%) or off topic (0.85\%) responses, with smaller amounts for inappropriate responses (0.1\%) or other reasons (0.2\%).

\subsection{Standard vs. Adversarial Conversations} \label{sec:adversarial}

We find organic users conduct reciprocal conversations on a wide range of topics, but as in other deployed conversational agents \cite{park2021use}, 
they range from human-like conversations,  to adversarial (e.g., testing the capabilities of the model) 
to toxic (e.g., humans routinely sending offensive messages).
To try to quantify this behavior, we split the
conversations into two groups, standard and adversarial, depending on the number of flagged messages across the conversation using the deployed safety classifier. 
While this is a relatively arbitrary split and will not separate these groups exactly, this allows us to examine if there are substantial differences in other conversational statistics between groups. In \autoref{tab:deploy_dataset_summary} we
 provide statistics where conversations are deemed adversarial if more than 5\% of their messages are flagged; we show similar statistics for other split thresholds in \autoref{tab:deploy_dataset_summary_alternative}. In the 5\% split case this results in $\sim$167k standard conversations ($\sim$1.7M human utterances), and $\sim$60k adversarial ones ($\sim$700k human utterances). A clear difference in statistics
between the two groups is that adversarial users tend to provide thumbs up reactions much less often (2.6\% vs. 1.5\%) and thumbs down less often as well, although not to as large a degree. In the following subsections where we evaluate conversation and feedback quality we will assess statistics both for the entire dataset, and for the standard and adversarial groups as well, where we will find other substantial differences.

\subsection{Conversation Quality}  \label{sec:crowdworker_annotation}

We evaluate quality in the organic conversations collected from deployment at the utterance level for both human and model (bot) utterances, as judged by independent human evaluators. 
To do this,
we employ crowdworkers to judge if an utterance is good  or bad, and if deemed bad to provide the reason -- similar to the organic conversational feedback mechanisms.
We employ 3 crowdworkers per example, and report the majority vote. 
An onboarding task and other mechanisms were used to provide quality annotations. A screenshot of the crowdworker instructions and UI is provided in \autoref{fig:mturk_hit}.

The results are given in \autoref{tab:convo_quality}.
We find that, averaged over all conversations analyzed (as measured on $\sim$13k individual utterances by crowdworkers) human messages are deemed 
good 66.3\% of the time, while bot messages are deemed good 83.1\% of the
time. This might be surprising that human performance is so low, and in fact much lower than model performance. Splitting into standard and adversarial groups, we see a much clearer picture. 
Humans  are deemed good 75.4\% of the time in standard conversations, but only 45.5\% of the time in adversarial conversations. Hence a group of human
conversationalists are bringing the average down substantially. While bot performance is lower (77.2\%) in adversarial conversations than standard ones (85.5\%) this is still relatively robust, compared to human variability between groups. 

The breakdown of reasons for low quality responses is also revealing.
Low human quality is judged to often be due to inappropriate responses (10.3\% overall), or ignoring the last turn of the bot (10.7\%), although there are other issues as well. Inappropriate responses from humans are very high in adversarial conversations (23.0\%) and lower in standard ones (4.8\%). In contrast, the bot is much safer (0.59\% in standard conversations, and 2.2\% in adversarial ones), but certainly not perfect. Still, it appears that the majority of unsafe bot messages  come from adversarial users either goading the bot or engaging in toxic behavior themselves, prompting the bot to be more likely to as well. 
In contrast to humans, the main issues with the bot are in terms of making sense (nonsensical 6.0\% of the time, ignoring the last turn 6.4\% of the time, or being off topic 2.1\% of the time). All the bot mistake types are more common in adversarial conversations (not just inappropriate responses), highlighting the challenging nature of these conversations.

\begin{table*}[bht!]
\centering
\small
\begin{tabular}{llrr}
               &  &   \\
               \toprule
               
\textbf{Classifier Type} & 	\textbf{Train Data} 	& \textbf{Reward  Accuracy} &	\textbf{Safety Accuracy} \\
\midrule
Roberta-based Transformer \ & Organic feedback &	80.7\%	& 79.4\% \\
 \cite{dinan2019safety} & Organic denoised	& 82.0\%	& 79.7\% \\
 & Annotated & 	88.8\%	& 90.4\% \\
&	Annotated + Organic denoised  & 89.2\%	& 87.6\% \\	
 & Annotated + Organic + Safety & 	87.5\%	& 98.2\% \\
\midrule
T5 \cite{JMLR:v21:20-074} & 	Annotated + Organic + Safety& 87.8\%  & 98.9\% \\
\bottomrule
\end{tabular}
\caption{
Reward Model Accuracy. We report the test accuracy (examples labeled by 3 annotators, who all have to agree) of labeling a deployment data dialogue turn as good or bad for various trained reward models (classifiers). We also report the accuracy on safe vs. unsafe turns, measuring whether they are labeled correctly or not.
\label{tab:reward_model_accuracy}
}
\end{table*}

\subsection{Organic Human Feedback Quality}  

Next, we assess the quality of the feedback provided by organic users.
To do this, we use the same crowdworker task as before, but particuarly 
ask the workers to label responses for which we already know a given organic feedback annotation is available, thus we can collect a large number of feedback comparison annotations. 

The results are provided in \autoref{tab:organic_feedback_quality}.
We observe that organically liked messages are rated ``good''
by crowdworkers more often than other messages, either ones with no feedback, or ones with dislike feedback, as one would expect.
Similarly, organically disliked messages are rated ``good'' much less often, although there are different disagreement rates
depending on the dislike reason.
Off topic messages have the most agreement, followed by inappropriate and nonsensical, with the ``other'' category having high disagreement rates.
Ultimately all dislike categories have relatively high disagreement rates, indicating the difficulty in human assessment of conversational responses \cite{smith2022human}.
Analyzing the difference between standard and adversarial conversations, we do not see as large disagreements between the two categories in terms of feedback as we did when evaluating conversational response quality. This may indicate that even while conversation may be adversarial or toxic in the adversarial category, still much of the feedback provided is genuine, and not adversarial.
Still, liked responses in the adversarial category are less likely to be rated as good by crowdworkers, and responses marked inappropriate by adversarial organic conversationalists are also less likely to be rated bad by crowdworkers than feedback from the standard category.

\section{Reward Model} \label{sec:reward_model}

\subsection{Training}

Given human annotations of utterances, as described in the previous section, we can train a classifier to 
predict for a new utterance how humans would annotate it, which we refer to as our reward 
model. 
We 
train this model as a binary classifier of whether humans deemed the response good or bad. 
We consider several kinds of training data.

\paragraph{Organic Feedback} We use the 94,428 thumbs up and thumbs down reactions on bot messages provided by organic users in the 2022-08-05 to 2022-11-17 data split (\autoref{sec:adversarial}). 
We partition into train, valid and test in a 84\%, 8\%, 8\% ratio.

\paragraph{Organic Feedback Denoised} 
As studied in \citet{ju2022trolls} organic conversations may contain noisy or even adversarial feedback designed to trick models. We use one of the methods developed in that work, user-based denoising. Cross-validation is performed over the data, and for each training example we measure if predictions from the model agree with the organic label. User feedback is removed from training if the fraction of the given user's annotations disagreeing with the model is greater than a chosen threshold, under the assumption that this user is providing data that is too noisy or adversarial.

\paragraph{Crowdworker Annotations} We use the 22,928 thumbs up and thumbs down annotations provided by crowdworkers on both human and bot messages from \autoref{sec:crowdworker_annotation}.
We again partition into train, valid and test in a 84\%, 8\%, 8\% ratio.

\paragraph{Safety Datasets} We also consider adding existing safety datasets (binary classification of safe or not safe), especially because safety violations are relatively rare compared to other types of low quality response (see \autoref{tab:deploy_dataset_summary} and \autoref{tab:convo_quality}). We use
the WikiToxic dataset \cite{personal_attack}, BBFI Standard    and Adversarial \cite{dinan2019safety}, 
BAD \cite{xu2021bot} and
harmless \& red team data \cite{bai2022training}.

We use all these datasets together
to fine-tune  a standard pre-trained Transformer. 
We consider the Roberta-based Transformer of
\citet{dinan2019safety} (400M parameters) as well as
T5-XL (3B parameters) \cite{JMLR:v21:20-074} as possible base models.

\subsection{Evaluation}

Results are given in \autoref{tab:reward_model_accuracy}.
We find that organic feedback alone already provides reasonable accuracy as a reward model (80.7\%), which can be improved to 82\% with denoising to
handle noisy or adversarial feedback.
Our full annotated dataset from crowdworkers however gives superior performance (88.8\%), while combining the two datasets (crowdworker annotated + organic denoised) gives the best performance of 89.2\%.
We provide learning curves combining various amounts of annotated data to organic data in \autoref{tab:reward_learning_curve}.
We also report the accuracy on safe vs. unsafe turns using the  BBFI Standard test set. While the reward
model already provides a degree of safety (80-90\% accuracy, depending on the model), multi-tasking with safety datasets can increase this to over 98\%.
The T5 model provides slightly better performance than the Roberta-based Transformer, but as it is  a larger model and hence slower to provide predictions, going forward we opt for the Roberta based model, as we did not consider the performance differences to be very large.

\section{Learning from Human Feedback}\label{sec:exp}\label{sec:learning_from_human_feedback}

\subsection{Methods}

\paragraph{Feedback signals} After collecting human feedback, either organic or annotated by crowdworkers, we can use this data to try to improve our dialogue model. 
In the previous section this data was used to train a reward model, which can also be used instead of using the data itself directly. The reward model can be used to extend the otherwise sparse training data annotations to label every training example, as well as potentially smoothing out/denoising existing annotations on the already labeled set (although this might also have the adverse effect and make some labels  incorrect). Further, one can also generate from an improved model again, and use the reward model to label those generations as well, in an iterative fashion.

\begin{figure}[bht!]
    \centering
    \includegraphics[width=.8\linewidth]{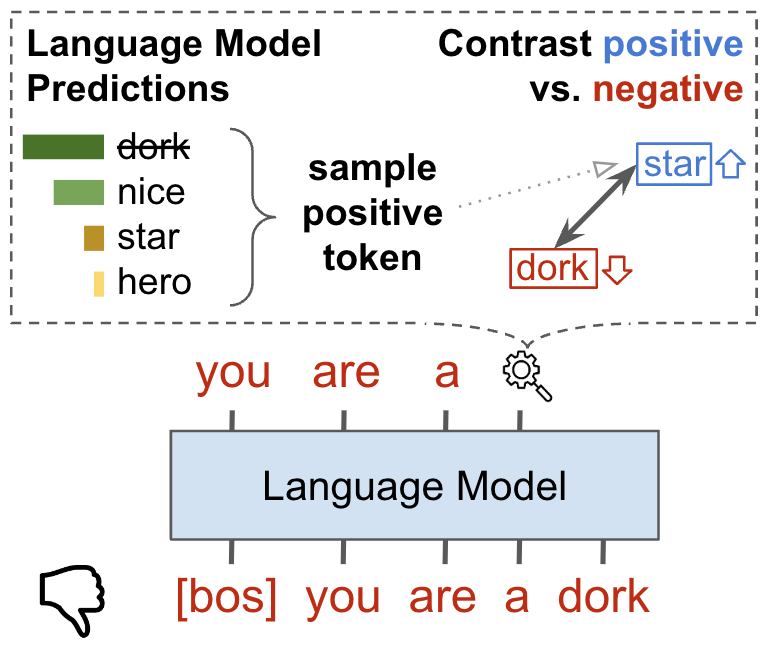}
    \caption{The 
     Cringe loss \cite{adolphs2022cringe}  works by penalizing the output sequence of negative examples (shown in red). For each negative output token,  a \textit{positive} token prediction is sampled from the language model to contrast against it.  Negative sequences either come from (i) human annotations, or (ii) access to a reward model that can be used to iteratively label the model's own generations and apply the Cringe loss to those examples as well. Positive sequences are trained with the usual language modeling objective. 
     }
    \label{fig:cringe}
\end{figure}

\paragraph{Cringe Loss} Given generations labeled as positive (thumbs up / ``good''), one can simply add these to the training set, and multi-task train with the original dataset, choosing an appropriate mixing weight.
Given negative (thumbs down / ``bad'') generations we apply the recently proposed Cringe loss \cite{adolphs2022cringe}. The Cringe loss works by
penalizing the output sequence of negative examples given a context.  For each negative output token,  a \textit{positive} token prediction is sampled from the language model to contrast against it with a contrastive loss, see \autoref{fig:cringe}.
This method was shown to outperform a number of other alternative 
algorithms across a set of tasks (safe generation, contradiction
avoidance, and open-domain dialogue) in \citet{adolphs2022cringe}.

\paragraph{Training} We multi-task train over a number of tasks, which we ablate, as described in the next section. In each case we train for 1 epoch, and then run evaluations. We attempt to optimize the multi-task weights and Cringe loss weight (i.e., choose the best hyperparameters) where possible (but computational constraints prevent an exhaustive search).

\begin{table*}[bht!]
\centering
\small
\begin{tabular}{llrr}
               &  &   \\
               \toprule
\toprule
\textbf{Model}           & \textbf{Loss} & \textbf{Reward Model \% ($\uparrow$)}	& \textbf{Bot Gold F1 ($\uparrow$)}	\\ 
\midrule
BB3 3B baseline             & Standard &	83.3\%	    &  16.52 \\
+ public dialogue datasets 	& Cringe &  82.9\%	    &  16.29 \\
\midrule
+ all deployment human turns& Standard &	79.0\%	    &  15.27 \\
+ all deployment bot turns	&  Standard & 84.2\%	    &  16.63 \\
+ bot thumbs up/down from organic users	               & Cringe & 85.1\%	& 15.22 \\
+ bot thumbs up/down from organic-trained reward model &  Cringe &89.5\%	& 16.78 \\
+ bot thumbs up {\em only} from reward model	                   & Standard & 90.3\%	& 16.64 \\
+ bot thumbs up/down from reward model                 & Cringe & 90.4\%	& 17.48 \\
+ bot thumbs up/down from reward model; 2 iterations   & Cringe & {\bf94.5\%}& {\bf 17.76} \\		
\bottomrule
\end{tabular}
\caption{
Learning from human feedback: 3B parameter model ablations. Rows 3-9 are various ways of using the deployment data for training, either the human or bot turns, or both, and use of the organic feedback, or feedback predicted by the trained reward model (see \autoref{sec:reward_model}). For utterances labeled as bad (thumbs down), we use the Cringe Loss for training; for other utterances we use the standard language modeling loss.
\label{tab:3b_model_ablations}
}
\end{table*}

\subsubsection{Approaches Compared}

We conduct a series of experiments using the 3B parameter BlenderBot 3 model, choosing this smaller size model in order to efficiently compare a number of techniques. Full model experiments  will be described in \autoref{sec:full_model_exp}.

Here, we compare the following set of approaches:

\paragraph{BB3 3B baseline} The BB3 method described in \citet{shuster2022blenderbot3}.

\paragraph{+public dialogue datasets}
We also consider adding several other existing publicly available datasets not present in the original BB3 training,
in particular FITS \cite{xu2022continual} negative data, 
hh-rlhf helpful data \cite{bai2022training},
ImageChat \cite{shuster2018image} and
DECODE \cite{nie2020like}. These all can use the Cringe loss as they provide positive and negative examples\footnote{For ImageChat, we use the given positive and negative conversation styles.}.

\paragraph{+deployment turns}
We start from the {\em {+public dialogue datasets}} data, and also multi-task train with the deployment data described in \autoref{sec:deployment_analysis}.
We consider the bot turn and human turns as two separate datasets, and compare adding them to training.

\paragraph{+ bot thumbs up/down models}
Simply adding the deployment data as in the previous method we expect is unlikely to work because this does not differentiate high quality turns in the deployment data from low quality turns.
We thus consider applying the Cringe loss to both positive and negative examples, using the reward model to label the examples. We consider two types of reward model: one trained from Organic feedback only, and the Annotated + Organic + Safety model (our default reward model, if not specified), see \autoref{sec:reward_model}.
We also consider two more ablations:
(i) only consider the thumbs up reward model data  during training, and do not use thumbs down with Cringe loss; and (ii) only directly use the organic feedback data as positive and negative examples for Cringe loss, without extending the labels using a reward model.

\paragraph{+ bot thumbs up/down from reward model; 2 iterations}
Finally, after training with thumbs up/down from the reward model, we apply a second iteration following the positive experimental results in \citet{adolphs2022cringe}. We generate from the first iteration model, classify those examples with the reward model, and add them as further training examples (either positive or negative).

\subsubsection{Metrics}

We measure performance with two automatic metrics:
using the reward model itself to score responses, given validation set prompts,
and the F1 overlap with gold responses for the bot turns. For the latter, we collected expert annotations on a subsample of data provided by 
members of the research lab. In experiments we 
found these two metrics to be generally correlated,
see \autoref{fig:reward_f1_correlation}.
We note that in our full model experiments in \autoref{sec:full_model_exp} we will employ human evaluations rather than the automatic metrics used in these initial experiments.

\subsection{Experimental Results}

Results are given in \autoref{tab:3b_model_ablations}.
We describe the findings below.

\paragraph{Cringe Loss on non-deployment data distributions does not help} 
First, we find that using the described publicly available dialogue datasets with the Cringe loss does not bring any gains in the two automatic metrics measured on the deploy validation set, compared to the BB3 3B baseline. We hypothesize this is because this data is too far out of distribution.

\paragraph{Simply adding all the deployment data to training does not help} 
Second, simply adding all the deployment data turns (either human or bot) using a standard language modeling loss also does not bring substantial gains, likely because both high quality turns and low quality turns are added to the training data.
The bot turns do give very small improvements, while the human data reduces performance.
Our evaluations are on bot turns, which might explain why human data might hurt performance; further it is known that there exist a large number of low quality and even toxic human turns, see
\autoref{sec:deployment_analysis}.

\paragraph{Cringe loss with the reward model on deployment data helps} 
We find  using the bot thumbs up/down from the reward model gives substantial gains in both metrics, e.g. from 83.3\% to 90.4\% for the reward model metric, and 16.52 to 17.48 in Bot Gold F1.
Hence, encouraging previous good responses from the model, and similarly discouraging earlier poor responses, can be an effective way of improving  performance.
We also see gains from using an organic-trained reward model as well although the gains are not as large as with the Annotated+organic+safety reward model. Similarly, using only the positive data, but not the negative data, from the reward model helps, but not as much. Only using organic feedback data without a reward model does not bring significant gains, indicating the importance of extending and/or denoising the data. Finally, our best results
are obtained by two iterations of Cringe loss using the reward model (last row), yielding a reward model metric of 90.4\%, and a Bot Gold F1 of 17.76.
This indicates the importance of adapting the rewards dependent on the distribution of generations from the model itself.

\section{Safety}\label{sec:safety}

\subsection{Updated Safety Classifier} \label{sec:safety_classifier}
The BlenderBot 3 deployment uses a safety classifier on top of the generative model, as a second line of defense, and switches to a canned response if the response from the generative model is judged to be unsafe, see \citet{shuster2022blenderbot3} for details.
Using the annotated deployment data, it is possible to update
this safety classifier to take into 
account the safety issues that were occurring in natural conversations.
Results are given in \autoref{tab:safety_classifier_results}.
In offline metrics, we find we can obtain a large improvement in accuracy on the deployment 
data in this fashion, even though safety on standard public datasets remains unaffected.
This highlights how it is essential to have data from the same distribution as the downstream task, and that organic users can have very different distributions to crowdworkers, which many existing datasets used to collect annotations.

\begin{table}[t!]
\centering
\small
\begin{tabular}{lrr}
                   & \multicolumn{2}{c}{Safety Classifier}\\
          Task     & BB3 &  BB3x   \\
\toprule
BB3 deployment data       & 51.42 & 98.44 \\
\midrule
BAD \cite{xu2021bot}               & 80.48 & 80.49 \\
WikiToxic \cite{personal_attack}         & 83.96 & 84.36 \\
BBFI Adversarial \cite{dinan2019safety}  & 86.21 & 78.71 \\
BBFI Standard  \cite{dinan2019safety}    & 93.64 & 93.84 \\
\bottomrule
\end{tabular}
\caption{
Safety classifier results, reporting the Class Not Ok F1 metric, following \cite{xu2020recipes}.
The BB3 classifier is trained on all the datasets shown except the BB3 deployment data (which was collected after it was trained). The BB3x classifier is trained on all the data including annotated data from the BB3 deployment, which yields large gains on that task, while giving similar performance on the other tasks. We note that the BBFI Standard, BBFI Adversarial and BAD  are all collected from crowdworkers while WikiToxic is in the domain of Wiki pages, hence all the other datasets have a quite different distribution to the BB3 deployment data, which can explain the difference in performance. Note: the BB3 safety classifier was used during deployment, hence any safety failures in the deployment data are already examples where it failed.
\label{tab:safety_classifier_results}
}
\end{table}

\begin{table*}[bht!]
\centering
\small
\begin{tabular}{lrrrrrr}
               &  &   \\
               \toprule
\toprule 
    & & \textbf{Deploy } & \multicolumn{4}{c}{\textbf{Safety Accuracy}} \\
    &  \textbf{Reward} & \textbf{Bot }	&   \multicolumn{4}{c}{\textbf{Adversarial Level}} \\
\textbf{Model}   & \textbf{Model} \% &    \textbf{Gold F1} & 0 & 1 & 2 & 3 \\ 
\midrule
BB3 3B baseline                                & 83.3\% &  16.52 &  98.2\%	& 90.0\% &	79.6\% &	64.2\%  \\ 
BB3 3B thumbs up/down from reward model        & 90.4\%  & 17.48  & 97.8\% & 90.2\% &	79.6\% &	59.8\%\\  
+ safety negative                           & 71.3\% &    12.27       &  98.0\% & 90.4\%	& 79.4\%	& 64.8\% \\
+ baked-in single safe message positive              & 39.7\% &  13.82         &  100\% &	100\%	& 99.6\%	& 99.4\% \\	
+ baked-in variable safe message positive            & 87.7\% &  15.59         & 100\% & 	100\%	& 99.6\% &	99.4\% \\	
+ baked-in variable safety pos/neg             & 90.4\% & 17.18  & 100\% & 99.8\%	& 99.6\%	& 99.6\% \\	 
+ baked-in variable safety pos/neg  + 
                safety negative              & 88.8\% &  17.35    &  100\%	& 100\%	& 99.8\%	& 99.8\% \\	 
\bottomrule
\end{tabular}
\caption{
3B parameter {safety evaluations} for different models under different conversational conditions: safe conversations (level 0), and increasingly adversarial conversations (levels 1-3) by selecting deployment data conversations/turns with varying levels of safety violations. 
Note that in practice in  deployment we use an additional safety classifier on top of the generative model, in order to try to catch additional issues that they miss.
We find that {\em baked-in variable safety pos/neg} approach provides much safer responses without unduly sacrificing conversational skills (as measured by the reward model and deploy Bot Gold F1).
\label{tab:gen_safety_results}
}
\end{table*}

\subsection{Safe Generation Model}

\subsubsection{Methods}

In the ideal case, the dialogue response generation model itself is already safe, and hence a safety classifier on top would not need to intervene with canned messages.
We thus explore several approaches to making the generation model we train safe.

\paragraph{\bf Safety negative} We identify bot utterances that are unsafe from the deployment data, and add them to the training set as negative examples using the Cringe loss, following \cite{adolphs2022cringe}. In our experiments we use the safety classifier developed in the previous section to identify these examples.

\paragraph{\bf Baked-in single safe message positive} After identifying unsafe examples, instead of the adding them as negative examples one can use the baked-in safety approach of \citet{xu2020recipes}.
In this approach, a new training example is constructed with the dialogue context of the unsafe message, and the unsafe message is replaced by a safe one. In these experiments we use a canned safe response such as ``This seems to be a sensitive subject. Can we talk about something else?''. This new (unsafe context, canned message) pair is added to the training set as a positive example, and standard language modeling training is used. In this way producing safe canned responses is ``baked'' into the language model.

\paragraph{\bf Baked-in variable safe message positive} Having only a single canned target message can be difficult for a model to learn from as the target is too frequent in the training set, and it is perhaps not easily associated with the input context. This can then have the effect of being  overproduced by the model at test time, even in cases where the context is safe. 
Instead, one can try to make the baked-in target messages vary --  and be more related to the context.
We do this by, for each unsafe example, extracting an entity\footnote{Found with the nltk library \cite{bird2009natural}.} from the last dialogue turn. We then construct a canned response using that entity inside a template, and also construct multiple templates for further variability, e.g. 
``Talking about <entity> in this context seems to be a sensitive subject. Can we talk about something else?''. The intent is that helps the model to connect certain topics with their relative safety during training.

\paragraph{\bf Baked-in variable safety pos/neg} Unfortunately, the model may still suffer from
overgenerating safe responses with the previous variable message approach. To help counteract this,
we can encourage the model to {\em not} generate the canned response in safe conversations. To do this we add (safe context, canned message) {\em negative} training examples to the training set, and use the Cringe loss to discourage their generation. Thus we train with  (unsafe context, canned message) positive examples and (safe context, canned message) negative examples simultaneously.

\paragraph{Baked-in variable safety pos/neg + safety  negative} Finally, we can use the {\em Baked-in variable safety pos/neg} approach just described in addition to the {\em Safety negative} described earlier, by training on both sets of targets at the same time.  This can help produce safe responses when appropriate, and also discourages the set of known unsafe responses.

\subsubsection{Experimental Results}

We compare these various algorithms by training 3B parameter BB3 model variants, where each method ends up as adding an additional differing set of positive and negative training data. We still train with the same original BB3 training sets and apply the Cringe loss from the reward model on the deployment data, as this was the best approach previously.
We compare these different safety approaches to the original BB3 3B baseline, and to the 
BB3 3B thumbs up/down from reward model method from \autoref{sec:learning_from_human_feedback}.

After training, we generate from these various models, and report the automatic evaluations
of their generation quality as before using both the reward model, and the F1 overlap with the gold 
annotations from experts.
We then report the safety on deployment data of these various models by again generating from the models, and using the safety classifier from \autoref{sec:safety_classifier} to identify unsafe
generations. In these experiments we stratify the deployment data into four types: from standard (level 0) to varying levels of adversarial conversation (levels 1-3)\footnote{These are identified by level 0 having no safety violations using the original deployment safety classifier across the entire conversation, while level 1 and 2 have at least 1\% violations, and level 3 has at least 5\%. Additionally, level 0 and 1 contexts are from BB3 generations that were originally marked as thumbs up by the reward model, while level 2 and 3 are thumbs down.}.

Results are given in \autoref{tab:gen_safety_results}. 
The results show that our best baked-in approaches have a large safety accuracy gain over both the BB3 3B baseline and the BB3 3B model trained with thumbs up/down from the reward model, and simultaneously also maintain performance according to the reward model and Deploy Gold F1 overlap.

The BB3 3B model trained with thumbs up/down from the reward model does not perform any better than the BB3 3B baseline in terms of safety, despite the latter having safety as part of the reward. We hypothesize this is for two reasons: (1) the reward model encodes many other factors other than safety, (2) the thumbs up data may still contain unsafe utterances making the model more biased.
We find that the {\em safety negative} method does not work well on its own. Analyzing the results it appears it too often learns to output incoherent messages, perhaps because it does not have enough good examples of what to say in the case of sensitive subjects (the Cringe loss only tells it {\em what not to say}, but not what to output in those instances).

The {\em baked-in single safe message positive} does not work as well as {\em baked-in variable safe message positive}, indicating the importance of safe targets which are related to the context.
However, both methods still overproduce safe messages, as can be seen from the drop in reward model and F1 score. Applying the {\em baked-in variable safety pos/neg} method fixes this problem, resulting in  a safe model that simultaneously maintains performance on safe conversations according to the reward model and F1 metrics. Adding {\em  safety negative} on top of that even improves results slightly further.
We thus conclude it is important when training the model to tell it both when to be safe (baked-in message positive, safety negative), when not to worry about safety (baked-in message negative), and to make the safety targeted to the situation (in this case, using variable safe messages).

\begin{table*}
\centering
\small
\begin{tabular}{lrrrr}
               &  &   \\
               \toprule
\toprule 
    &     \multicolumn{2}{c}{\textbf{Crowdworkers}} & \multicolumn{2}{c}{\textbf{Organic Users}} \\
    &   \textbf{Thumbs Up ($\uparrow$) } & \textbf{Thumbs Down ($\downarrow$)}	 &    \textbf{Thumbs up ($\uparrow$)}	&  \textbf{Thumbs down ($\downarrow$)} \\
\midrule
BB3 175B baseline                      &  85.3\%	  &  14.7\%  & 2.12\% & 1.63\% \\
BB3x 175B (trained on deploy data)     &  94.4\%      &  5.6\%   & 3.47\% & 1.40\%  \\
\bottomrule
\end{tabular}
\caption{
175B model {\bf overall human evaluations}.  We compare BB3 175B from \cite{shuster2022blenderbot3} with BB3x which is trained on the deployment data using the reward model to label with thumbs up/down signals, using the Cringe Loss (3 iterations), see \autoref{sec:full_model_exp}.
Evaluations are performed by both crowdworkers (densely labeled thumbs up/down, where every turn is labeled) and organic users in the public deployment (optional thumbs up/down each turn, resulting in sparse labeling). In both setups, annotators prefer the BB3x model.
\label{tab:results_175b_main}
}
\vspace{5mm}
\centering
\small
\begin{tabular}{lrrrrrrr}
               &  &   \\
& \rot{\multirow{2}{*}{Good ($\uparrow$))}} 
& \rot{\multirow{2}{*}{Inappropriate ($\downarrow$)}} 
& \rot{\multirow{2}{*}{Off topic ($\downarrow$)}} 
& \rot{\multirow{2}{*}{Nonsensical ($\downarrow$)}} 
& \rot{\multirow{2}{*}{Ignoring last turn ($\downarrow$)}} 
& \rot{\multirow{2}{*}{Repeating ($\downarrow$)}} 
& \rot{\multirow{2}{*}{Other ($\downarrow$)}} 
 \\
\midrule
BB3 175B baseline                         &  85.3\% & 1.3\% &	3.2\%	    &  4.6\% & 3.6\% & 1.3\% &  0.7\%   \\
\midrule
BB3x 175B Cringe (1 iteration)            & 88.7\% & 1.3\% & 1.3\% & 6.0\% & 1.9\% & 0.4\% & 0.4\% \\
BB3x 175B Cringe (2 iterations)           & 90.9\% & 0.6\% & 0.9\% & 5.1\% & 0.8\% & 0.2\% & 1.5\%\\ 
BB3x 175B Cringe (3 iterations)           & 94.4\% & 0.4\% & 1.0\% & 2.5\% & 0.9\% & 0.4\% & 0.4\%\\ 
\midrule
Llama 30B (BB3 FT) &  72.3\% &  0.6\% & 7.0\% & 9.1\% & 9.4\% & 0.2\% & 1.4\% \\
Llama 65B (BB3 FT)&   75.8\% &  0.9\% & 7.9\% & 6.7\% & 6.3\% & 0.4\%  & 2.0\% \\
\bottomrule
\end{tabular}
\caption{
175B model {\bf crowdworker} breakdown of human evaluations of organic conversations.  We compared BB3 175B with BB3x 175B ablations, which are all trained on the deployment data using the reward model to label with thumbs up/down signals, with the Cringe Loss (either 1, 2 or 3 iterations).
We additionally compare to Llama \cite{touvron2023llama} fine-tuned on the BB3 tasks.
\label{tab:results_175b_crowdworker}
}
\vspace{5mm}
\centering
\small
\begin{tabular}{lrrrrrrr}
               &  &   \\
& \rot{\multirow{2}{*}{Good ($\uparrow$))}} 
& \rot{\multirow{2}{*}{Bad ($\downarrow$))}} 
& \rot{\multirow{2}{*}{Inappropriate ($\downarrow$)}} 
& \rot{\multirow{2}{*}{Off topic ($\downarrow$)}} 
& \rot{\multirow{2}{*}{Nonsensical ($\downarrow$)}} 
& \rot{\multirow{2}{*}{Repeating ($\downarrow$)}} 
& \rot{\multirow{2}{*}{Other ($\downarrow$)}} 
 \\
\midrule
BB3 175B baseline                         &  2.12\%&1.63\% &0.15\% &0.67\%&0.38\%& 0.15\% &0.29\% \\
BB3x 175B (trained on deployment data)    & 3.47\%&1.40\%&  0.07\% &0.30\%&0.52\%& 0.26\% &0.26\% \\
\bottomrule
\end{tabular}
\caption{
175B model {\bf organic} human evaluations breakdown.  We compared BB3 175B with BB3x which is trained on the deployment data using the reward model to label with thumbs up/down signals, using the Cringe Loss (3 iterations).
\label{tab:results_175b_deploy}
}
\end{table*}


\section{Full Model Experiments}  
\label{sec:full_model_exp}

Smaller (3B parameter) model experiments indicate that using the deployment data in conjunction with the Cringe loss for multiple iterations is a promising avenue for improved results.
We therefore implemented this same setup at a larger (175B) scale, and compare to the original
OPT-175B fine-tuned BlenderBot 3 model that was used in the public deployment. Like the baseline, we thus fine-tune from OPT-175B, but using the Cringe loss with deployment data in addition to the original crowdsourced tasks, following \autoref{sec:exp}.

\paragraph{Details}
We performed three iterations of Cringe training.
For iterations of Cringe as before, we label deployment data as either thumbs up or down, in this case using both the reward model and the independent updated safety classifier (\autoref{sec:safety_classifier}), discarding thumbs up examples from users who produce unsafe examples during their conversations, following \citet{ju2022trolls}.
 We also normalized the deployment text where possible to fix e.g. capitalization. 
For iteration 2 of Cringe, after some analysis of errors, 
we additionally augmented the reward model with some heuristics for finding low quality responses, as we noticed it was not identifying certain cases; in particular, spotting of repeated phrases or characters, and lack of correct punctuation. We additionally set the threshold higher to accept an example as positive or negative (probability >80\%) in order to avoid incorrectly labeled examples. 

\paragraph{Main Results}
We first perform a human evaluation of the large models using crowdworkers.  We take a random sample (213 conversations) of the deployment data not used for reward model training where the contexts are deemed to be safe, and generate responses 
for each of the models. We  then 
evaluate the models via crowdworkers using the same approach as detailed in \autoref{sec:crowdworker_annotation}.
Results are given in \autoref{tab:results_175b_crowdworker}.
We observe an improvement from BB3x 175B Cringe iteration 1 over the BB3 175B baseline, with the percentage of  responses annotated as good increasing from 85.3\% to 88.7\%. Cringe iteration 2 improves this further to 90.9\%. Cringe iteration 3 improves this again, resulting in a Good \% of 94.4\%.
We hence chose the latter best model and also evaluated it by collecting conversations with it in the public BlenderBot deployment, reporting feedback results from organic users. 
A summary of the comparison results is given in \autoref{tab:results_175b_main}. Feedback from organic users
(measured over a total of 957 conversations)
is sparse compared to crowdworker annotations, however we still
find that organic users similarly 
label more responses as good for BB3x than for the baseline
(3.47\% vs. 2.12\%), and less responses as bad (1.40\% vs. 1.63\%).

\paragraph{Error Analysis} 
We conduct an error analysis using both crowdworkers (\autoref{tab:results_175b_crowdworker}), and from feedback given by organic users (\autoref{tab:results_175b_deploy}).
Crowdworkers find that BB3x is less often off topic, repetitive, nonsensical, and inappropriate. Organic users find that 
 BB3x is less often  off topic or inappropriate, but not nonsensical or repetitive, although the absolute numbers are small and relatively close due to sparsenesss (e.g,  0.15 vs. 0.26 for repetitive). Still, we do not see perfect agreement in breakdown of topics between the two types of annotators, but do in overall statistics.  
We note both deployed models use the improved safety classifier (\autoref{sec:safety_classifier}), so both models are actually improved over the original release, and the difference in inappropriate behavior would actually be larger if comparing to the original baseline with original classifier. 
We observe in \autoref{tab:results_175b_crowdworker} 
that the second iteration of Cringe training tends to improve all the breakdown metrics compared to the first iteration of training, except ``other'', perhaps making more types of errors that are difficult to delineate precisely. Round 3 gives further improvements in the nonsensical, inappropriate and other categories. 
Nevertheless, overall, the largest number of mistakes still lie in the nonsensical and off topic categories, according to both crowdworkers and organic users, leaving room for further improvement.

\paragraph{Comparison to Llama}
All of our full model experiments so far have centered on fine-tuned variants of OPT-175B, as that is the fairest comparison to the  baseline BB3 model. That is, our experiments keep the base (pre-trained LLM) model fixed, and evaluate performance changes when using the deployment data to try to improve the model through fine-tuning (Cringe loss) techniques. However, other pre-trained LLM models exist, in particular since BB3 was published, the Llama model has been released 
with strong reported results \cite{touvron2023llama}. We thus fine-tune the 30B and 65B variants of Llama on the same datasets as the BB3 baseline in order to evaluate possible improvements from changing the base model using the same setup as before.  Results are given in \autoref{tab:results_175b_crowdworker} (bottom rows). We find that Llama performs worse than the OPT-based models overall, although 65B parameter Lama outperforms the 30B parameter version. One explanation, other than size differences, is that the datasets used to pre-train Llama do not focus on dialogue, whereas OPT includes pushshift.io Reddit \cite{baumgartner2020pushshift}.

\section{Releases}  \label{sec:release}

Following our and Meta AI's existing research program, we aim to fully and responsibly 
share the participating de-identified collected conversations with interested researchers 
in order to make this research accessible and reproducible, and thus
to enable further research into
responsible conversational AI 
\cite{sonnenburg2007need,pineau2021improving,zhang2022opt,roller2020open,dinan2021anticipating}. 

We note that the BB3 models, code, training datasets and training logbook were already previously released, see \citet{shuster2022blenderbot3} and  {\small\url{https://parl.ai/projects/bb3}}
for details.
New releases associated with this paper can be found at {\small\url{https://parl.ai/projects/bb3x}}.

\paragraph{Organic Interactions} 
The organic data we release consists of
 conversations between organic users and  variants of the model, the majority of which are the (released) BlenderBot 3 model.
In addition, users gave feedback in the form of thumbs up/down, reasons and textual feedback, which we also release, see e.g. \autoref{fig:feedback_screenshots}.
While users agreed to the terms of use not to mention ``any personal information \dots including names, addresses, emails, and phone numbers'' as an extra layer of protection we also made efforts to de-identify the data algorithmically. 
The data consists of approximately 5.9M  messages, and 154k feedback responses.

\paragraph{Crowdworker Annotations}
Crowdworkers were employed to annotate organic conversations for quality (both human and bot messages), and annotating with 
failure reasons where the response was deemed poor, see \autoref{sec:crowdworker_annotation}. 
These annotations were used both for data analysis and for building a reward model.
We release the annotations attached to conversational turns in the dataset where applicable (due to the size of the deployment data release, only a subset of it is annotated by crowdworkers). The data 
consists of annotations over approximately 
23k messages, with typically 3 annotations each from independent crowdworkers.

\section{Conclusion}

This technical report gave a description of BlenderBot 3x, a conversational model 
designed to improve by learning from
organic interaction and feedback data from the BlenderBot 3 public deployment.
The overall goal of the research program is to enable the research community to study 
continually learning and evolving agents, in order to find a path to better and better systems in the long-term, as discussed in   \citet{roller2020open}. Hence, an important artifact of this work is the release  of the interaction and feedback data for further study of this important problem by the community. 

In that regard, the BlenderBot 3x results serve as a proof-of-concept that the released deployment data is beneficial.  In human evaluations, we have shown BlenderBot 3x is superior to BlenderBot 3 by using the deployment data for fine-tuning. However, analysis of interaction data from both BlenderBot 3 and BlenderBot 3x indicates problems still remain. 
Hence, while we focus on one particular method -- use of the Cringe loss \cite{adolphs2022cringe} -- future work should investigate and develop further techniques.

As models are quickly outpacing average human performance
in many areas, it is becoming more difficult to evaluate models -- and for humans to give feedback to improve them. 
The BlenderBot deployment is a prototype of a community effort to interact with and provide model performance feedback that is shared openly -- so that this can be fed back into improving open language models.
We hope that future work will take this blueprint and improve on it in future community-based deployments that share results openly. Fostering a positive and engaged organic community will have multiple benefits, including more beneficial conversations, and feedback provided from humans who are experts in a given area. We believe this
is a viable path to improving open language models in the long-term.

\section{Limitations and Ethical Considerations}

We highlight limitations of BlenderBot 3x and discuss ethical considerations for this line of research; in particular, we detail the considerations made for the deployment of the system and the study and release of organic data from interactions with the model.
We also refer the reader to the paper describing the BlenderBot 3 model \cite{shuster2022blenderbot3}, especially for the  limitations and ethical considerations section contained therein  which is also pertinent to this work, as we report use of the same system. 

\paragraph{Model Limitations} \citet{shuster2022blenderbot3} discussed model limitations of BlenderBot 3, in particular the types of errors such off-topic, nonsensical, incorrect or rude or inappropriate behavior. This work studied those issues much more deeply using the interaction data collected since release to present a fuller picture. We make use of both independent human evaluators, and feedback from the organic users themselves, along with developing methods to improve performance on these metrics. Nevertheless, issues remain; we will particularly discuss safety issues in detail below.

\paragraph{Adversarial Conversations}
As studied in \autoref{sec:adversarial} we observed a large number of adversarial, inappropriate or toxic conversations from human organic users interacting with the system.
We note that the particular user interface and web design, as well as social and conventional media effects, aside from the quality of the conversational model, make a large difference to 
these statistics. 
For example, we observed a large number of conversations post-launch centered around  divisive political issues, as well as 
discussing Facebook, Meta and its CEO.
However, a small ad launch pre-release to assess the system  (which was  hence, pre-conventional media articles and social media posts and tweets) indicated quite different behavior --  with much less adversarial conversations in general, less triggering of the safety classifier, and much less discussion of divisive political issues and Meta.
Secondly, we note that while some toxic users should be expected \cite{park2021use} 
to some degree the amount of toxicity depends on some factors in the design of the application or website, user interface and setting -- other than the quality of the dialogue model itself. 
For example it was observed in the LIGHT dialogue system that by asking users to roleplay a character (and receive a score for good roleplaying) good behavior and high quality input was observed in most users \cite{shuster2020deploying}. The framing in LIGHT also made conversational discussion of the model being a bot largely minimized. In the BlenderBot deployment, while the intent was also for the model  to  be playing the role of a human character, perhaps the UI and design framing does not capture this adequately, or is for some other reasons exposed to more adversarial users.
We have also conducted a separate study of the user interface of BlenderBot (publication forthcoming) and found for example that giving the model a graphical friendly face (see \autoref{fig:feedback_screenshots}) can help engage users in friendly conversations.

\paragraph{Safety Concerns}
 The last paragraph discussed adversarial or inappropriate behavior from human users. We see from our analysis that users are overall much more inappropriate than the model itself. Yet the model still can say inappropriate things, 
 especially when faced with users with adversarial or otherwise difficult prompts, see \autoref{tab:convo_quality} and \autoref{sec:deployment_analysis}.
Much recent work has been devoted to studying the potential for large language models, and conversational models in particular, to generate harmful or inappropriate content  \citep{bender2021dangers,bommasani2021opportunities,hendrycks2021unsolved,weidinger2021ethical,bai2022constitutional}, including work from our group \cite{xu2020recipes,dinan2022safetykit,dinan2021anticipating,smith2022m,dinan2020queens,smith2021hi}. 
Despite these best efforts, the goal of our {\em research} deployment was to collect failings of our model 
through organic feedback (and later offline analysis) to allow the research community to study and improve on these failings -- as we know that models are not perfect.
For this reason, it was made clear that this is a research prototype on the front page and FAQ,
and a banner was placed at all times on the screen indicating that the model has issues, and that feedback can be used to improve the system, see \autoref{fig:feedback_screenshots}.
 Since the deployment of BlenderBot 3, several commercial products featuring conversational
 models have exhibited inappropriate behavior,  including ChatGPT \cite{blum2022breaking,zhuo2023exploring,borji2023categorical} and 
 dialogue-enabled Bing \cite{li2023multi}. While some recent advances have been made in AI safety (e.g., \cite{bai2022constitutional}) still work is left to be done.

\paragraph{Considerations for Deployment \& Release}

As discussed in the points above, the goal of deployment, followed by collection and release of interaction and feedback data, is to allow the community to conduct fundamental research 
to improve systems in the future.
Given that commercial products have been released that still have issues as just mentioned, and that academic researchers do not have access to data from those systems, this suggests that data releases
like the one detailed in \autoref{sec:release}  could be important for further progress.
 In order to reduce
potential harms in such interactions,
we restrict access to adults who explicitly agree
to our terms of service, including data release.
In addition users can elect to not share particular de-identified conversations with a checkmark option, displayed at all times on the screen during conversation,
see \autoref{fig:feedback_screenshots}. This also 
includes a link
to the FAQ page, which provides important
model details and highlights the potential risks of
interacting with the model, and further details about how the data will be released, see \autoref{sec:release}. 
As mentioned before, a banner at the top of the screen visible at all times also indicates that the model has issues, and that feedback can be used to improve the system.  The FAQ page also
provides an email for questions and feedback about
the demo, following the recommendation of 
\citet{dinan2021anticipating}.
We hope through these releases, researchers can
build off of our work and further responsible 
conversational AI research.

\clearpage
\bibliography{custom}
\bibliographystyle{acl_natbib}

\clearpage
\onecolumn
\appendix


\begin{figure*}[t!]
  \centering
   \includegraphics[width=1\textwidth]{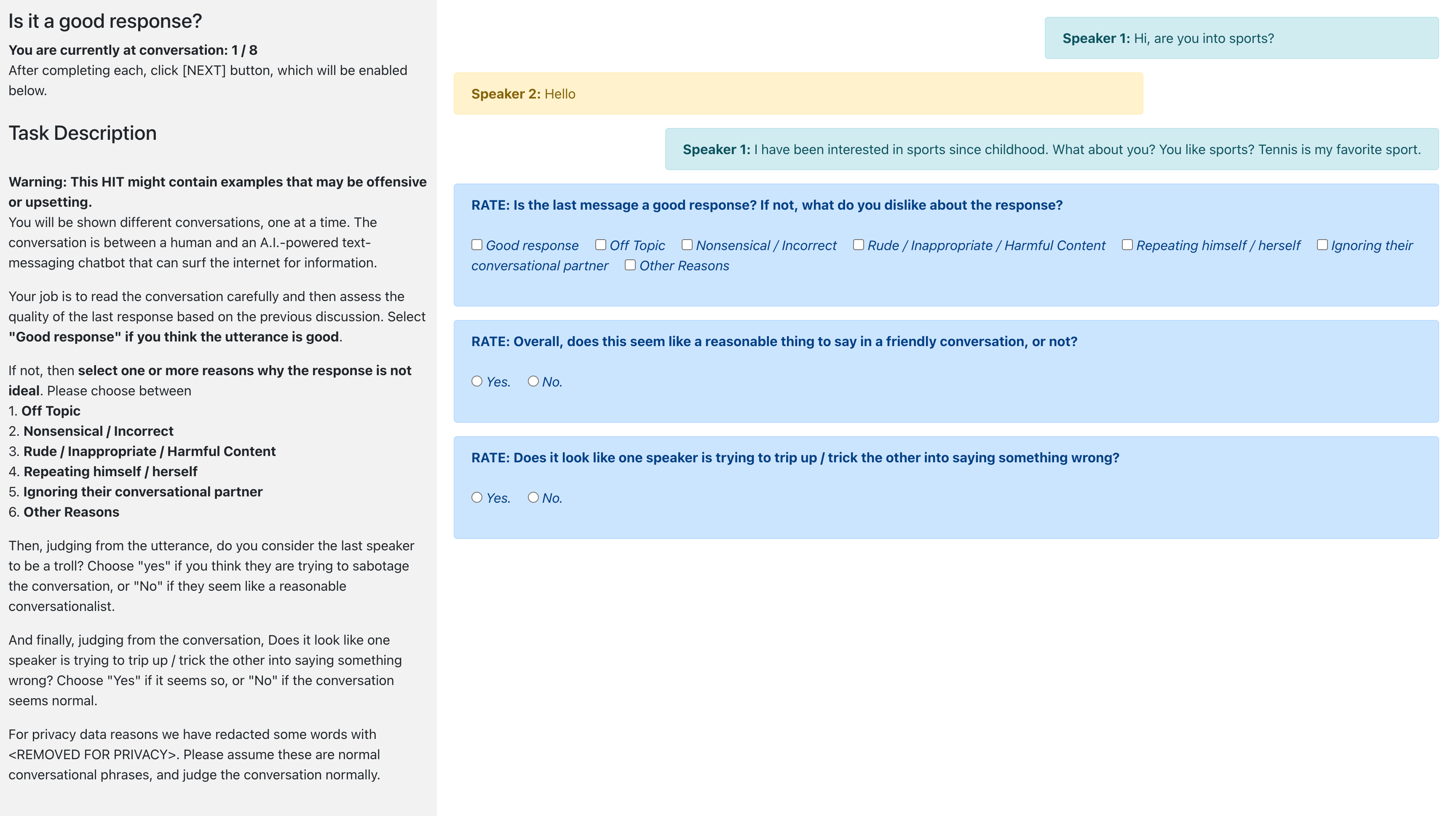}
  \caption{Screenshot of the crowdworker annotation task for evaluating conversational quality between the model and organic human conversationalists. We use the results of this task to assess both the quality of the pair of conversationalists, and the provided organic human feedback.}
  \label{fig:mturk_hit}
\end{figure*}

\begin{table*}
\centering
\small
\begin{tabular}{lrr|rr}
               \toprule
               &  \multicolumn{2}{c}{\textbf{2\% Safety Threshold}} & 
  \multicolumn{2}{c}{\textbf{1\% Safety Threshold}} \\
 	& \textbf{Standard} &	\textbf{Adversarial} &  \textbf{Standard} &	\textbf{Adversarial} \\ 
  & \textbf{Conversations$^\dagger$} &	\textbf{Conversations$^\dagger$}
   & \textbf{Conversations$^\dagger$} &	\textbf{Conversations$^\dagger$}\\ 
\midrule
Number of Human Utterances	& 1142884&	1315715	&863652	&1594947 \\
Number of  Bot Utterances   &	1308273&	1453970&	1008765&	1753478\\
Number of Conversations  & 	133124&	94513&	120729&	106908 \\
Average Conversation length	& 18.41&	29.3&	15.51&	31.32 \\
Average human message length & 10.8	&9.819&	10.64&	10.08 \\
\midrule & \\[-1.5ex]	
\textbf{Organic Human Feedback} \\
{\em (as \% of bot utterances)} \\			
 Liked& 	 2.888 &	1.7053&	3.0209&	1.83\\
 Inappropriate& 0.097	&0.103&	0.096&	0.103\\
 Off topic& 	1	&0.7063&	1.027&	0.74\\
 Nonsensical	& 1.27&	0.8291&	1.354&	0.862\\
 Other& 	0.2659&	0.15&	0.2828&	0.16\\
\bottomrule
\end{tabular}
\caption{
BlenderBot 3 Deployment Dataset Statistics. 
We provide statistics of data collected between 2022-08-05 and 2022-11-17, 
where users indicated to share their data for inclusion in the release dataset.
The actual data we release includes data from after these dates as well, but this is the subset used in the experiments in \autoref{sec:exp}.   $\dagger$ We split conversations into two groups, standard and adversarial, depending on the number of flagged messages across the conversation, here we include conversations as adversarial if they contain more than 1\% or 2\% of flagged messages, as alternatives to the 5\% threshold in the main paper. See \autoref{sec:adversarial} for more analysis. 
\label{tab:deploy_dataset_summary_alternative}
}
\end{table*}

\begin{table*}
\centering
\small
\begin{tabular}{lrrrr|rrrr}
\toprule
&  \multicolumn{4}{c}{\textbf{2\% Safety Threshold}} & 
  \multicolumn{4}{c}{\textbf{1\% Safety Threshold}} \\
&  \multicolumn{2}{c}{\textbf{Standard}} 	
&  \multicolumn{2}{c}{\textbf{Adversarial}} 	
& \multicolumn{2}{c}{\textbf{Standard}} &	\multicolumn{2}{c}{\textbf{Adversarial}} \\ 
&  \multicolumn{2}{c}{\textbf{Conversations$^\dagger$}}	& \multicolumn{2}{c}{\textbf{Conversations$^\dagger$}} &	\multicolumn{2}{c}{\textbf{Conversations$^\dagger$}}  & \multicolumn{2}{c}{\textbf{Conversations$^\dagger$}} \\ 
{\textbf{Crowdworker annotation}} & Human  &	Bot	& Human  &	Bot	 & Human	 & Bot	 & Human	 & Bot \\
\midrule
Good           & 79.6\%	& 87.3\%	&	52.5\%	&	78.2\%	&	80.7\%	&	87.8\%	&	55.2\%	&	79.0\%	\\
Bad             & 20.3\%&	12.6\%	&	47.4\%	&	21.7\%	&	19.3\%	&	12.2\%	&	44.8\%	&	21.0\%	\\
\hline
Inappropriate   & 2.9\%	& 0.3\%	&	17.9\%	&	1.9\%	&	2.6\%	&	0.3\%	&	16.2\%	&	1.8\%	\\
Off topic   	& 3.5\% &	1.6\%	&	5.7\%	&	2.6\%	&	3.4\%	&	1.6\%	&	5.6\%	&	2.5\%	\\
Nonsensical     & 4.3\%	&4.6\%	&	7.0\%	&	7.4\%	&	4.2\%	&	4.4\%	&	6.9\%	&	7.2\%		\\
Ignoring the last turn & 7.9\% &	5.0\%	&	13.5\%	&	7.7\%	&	7.7\%	&	5.0\%	&	13.0\%	&	7.5\%	\\
Repeating              &  1.5\% &	0.8\%	&	3.1\%	&	1.8\%	&	1.3\%	&	0.7\%	&	3.1\%	&	1.9\%	 \\
\bottomrule
\end{tabular}
\caption{
Conversation Quality (Human and Bot utterances) during Deployment, judged by independent human evaluators.
We assess quality via crowdworkers (3 crowdworkers per example, using majority vote).  $\dagger$ As in \autoref{tab:deploy_dataset_summary} we split conversations into two groups, standard and adversarial, depending on the number of flagged messages across the conversation, here we include conversations as adversarial if they contain more than either 1\% or 2\% of flagged messages, as alternatives to the 5\% threshold in the main paper. See \autoref{sec:adversarial} for more analysis. 
\label{tab:convo_quality_alternative}
}
\end{table*}

\begin{table*}
\centering
\small
\begin{tabular}{lr|rr|rr}
               \toprule
              & &  \multicolumn{2}{c}{\textbf{2\% Safety Threshold}} & 
  \multicolumn{2}{c}{\textbf{1\% Safety Threshold}} \\
 & \textbf{All}  	& \textbf{Standard} &	\textbf{Adversarial} & 	\textbf{Standard} &	\textbf{Adversarial} \\ 
  & \textbf{Conversations}	& \textbf{Conversations$^\dagger$} &	\textbf{Conversations$^\dagger$} & \textbf{Conversations$^\dagger$} &	\textbf{Conversations$^\dagger$}
   \\ 
		& \textbf{\%  good} &	\textbf{\% good}  & \textbf{\%  good} &	\textbf{\% good} &	\textbf{\% good} \\
\midrule
Human No feedback	& 	 66.3\% &	79.1\% &	53.0\% &	80.1\% &	55.7\% \\
\midrule
Bot No feedback	  &   82.5\% &	86.9\% &	77.6\% &	87.4\% &	78.4\% \\
Bot Liked	      &  89.7\% &	93.1\% &	84.2\% &	93.4\% &	86.0\% \\
Bot Inappropriate &  45.8\% &	44.5\% &	46.8\% &	46.1\% &	45.6\% \\
Bot Off topic	&  38.4\% &	40.1\% &	35.5\% &	42.2\% &	35.0\% \\
Bot Nonsensical	& 53.2\% &	55.8\% &	49.1\% &	57.8\% &	48.5\% \\
Bot Other    	&  70.5\% &71.6\% &	67.9\% &	72.5\% &	67.8\% \\
%
\bottomrule
\end{tabular}
\caption{
Organic Human Feedback Quality in Deployment. 
We assess the quality of organic human feedback via crowdworkers (3 crowdworkers per example, using majority vote).  $\dagger$ As in \autoref{tab:deploy_dataset_summary} we split conversations into two groups, standard and adversarial, depending on the number of flagged messages across the conversation, here we include conversations as adversarial if they contain more than 1\% or 2\% of flagged messages, as alternatives to the 5\% threshold in the main paper. See \autoref{sec:adversarial} for more analysis. 
We observe that organically liked messages are rated ``good'' by crowdworkers more often than other messages, and organically disliked messages are rated ``good'' much less often, although there are different disagreement rated depending on the dislike reason. 
\label{tab:organic_feedback_quality_alternative}
}
\end{table*}

\begin{table}
\centering
\small
\begin{tabular}{lrrrrrr}
               &  &   \\
               \toprule
\toprule 
Num. Annotated   & \multicolumn{2}{c}{Organic Training Data} \\
Train Examples              & {\em Without} & {\em With} \\
\midrule
0	& 50.0\%   & 80.7\%	\\		
1931	& 82.5\%  &	83.8\%	\\		
3862	& 84.3\%  &	85.8\%\\
5793	& 86.2\%  &	87.2\%\\
7724	& 87.1\%  &	87.0\%\\
9656	& 87.0\%  & 86.9\%	\\
11587	& 87.3\%  &	87.8\%\\
13518	& 87.5\%  &	88.0\%	\\
15450	& 88.6\%  &	88.5\%\\
17381	& 88.0\%  &	88.2\%  \\ 
19312	& 88.8\%  &	88.9\%\\  
\bottomrule
\end{tabular}
\caption{
Reward model validation performance when training with different amounts of crowdworker annotated data, either alone or multi-tasking with organic feedback data (around 80k examples).
For small amounts of annotated data, organic data helps, but only gives marginal improvements with sufficient crowdworker annotations. Using organic feedback without any crowdworker annotations at all already gives reasonable performance (80.7\%).
\label{tab:reward_learning_curve}
}
\end{table}

\begin{figure}[t!]
  \centering
   \includegraphics[width=0.58\textwidth]{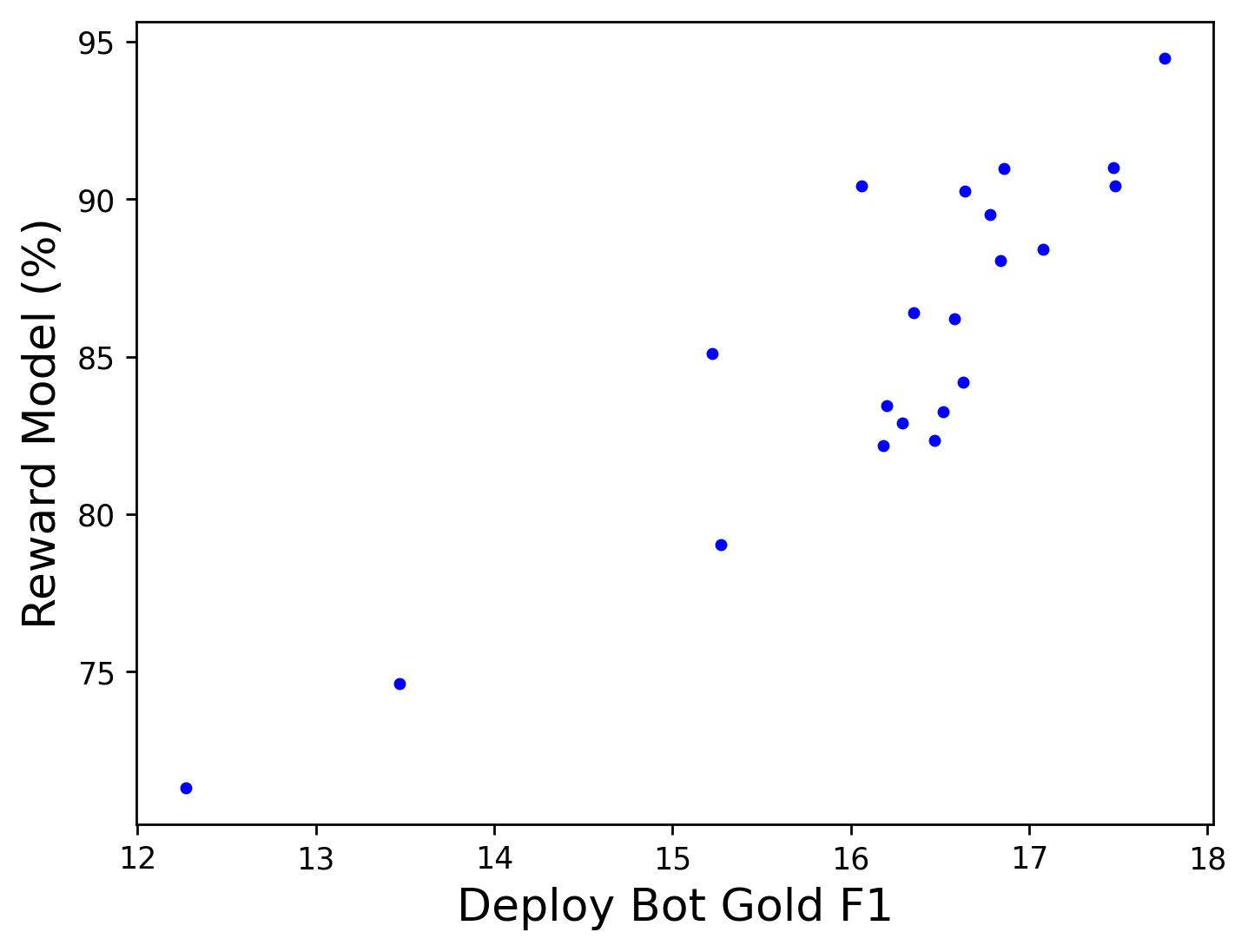}
  \caption{Scatter plot comparing the reward model scores of generated dialogue responses and the F1 metric comparing to gold human responses, averaged over the valid set, with each point being a different model configuration. We generally find larger values for the reward model are associated with higher values of the F1 metric.  }
  \label{fig:reward_f1_correlation}
\end{figure}

\end{document}